# Identifying Finite Mixtures of Nonparametric Product Distributions and Causal Inference of Confounders

**Eleni Sgouritsa**[1], **Dominik Janzing**[1], **Jonas Peters**[1,2], **Bernhard Schölkopf**[1]

[1] Max Planck Institute for Intelligent Systems, Tübingen, Germany
[2] Department of Mathematics, ETH, Zurich
{sgouritsa, janzing, peters, bs}@tuebingen.mpg.de

## Abstract

We propose a kernel method to identify finite mixtures of nonparametric product distributions. It is based on a Hilbert space embedding of the joint distribution. The rank of the constructed tensor is equal to the number of mixture components. We present an algorithm to recover the components by partitioning the data points into clusters such that the variables are jointly conditionally independent given the cluster. This method can be used to identify finite confounders.

## 1 Introduction

Latent variable models are widely used to model heterogeneity in populations. In the following (Sections 2-4), we assume that the observed variables are jointly conditionally independent given the latent class. For example, given a medical syndrome, different symptoms might be conditionally independent. We consider $d \geq 2$ continuous observed random variables $X_1$, $X_2$,...,$X_d$ with domains $\{\mathcal{X}_j\}_{1 \leq j \leq d}$ and assume that their joint probability distribution $P(X_1, \ldots, X_d)$ has a density with respect to the Lebesgue measure. We introduce a finite (i.e., that takes on values from a finite set) random variable $Z$ that represents the latent class with values in $\{z^{(1)}, \ldots, z^{(m)}\}$. Assuming $X_1, \ldots, X_d$ to be jointly conditionally independent given $Z$ (denoted by $X_1 \perp\!\!\!\perp X_2 \perp\!\!\!\perp \ldots \perp\!\!\!\perp X_d \,|\, Z$) implies the following decomposition into a finite mixture of product distributions:

$$P(X_1, \ldots, X_d) = \sum_{i=1}^{m} P(z^{(i)}) \prod_{j=1}^{d} P(X_j | z^{(i)}) \qquad (1)$$

where $P(z^{(i)}) = P(Z = z^{(i)}) \neq 0$.

By *parameter identifiability* of model (1), we refer to the question of when $P(X_1, \ldots, X_d)$ uniquely determines the following parameters: (a) the number of mixture components $m$, and (b) the distribution of each component $P(X_1, \ldots, X_d | z^{(i)})$ and the mixing weights $P(z^{(i)})$ up to permutations of $z$-values. In this paper, we focus on determining (a) and (b), when model (1) is identifiable. This can be further used to infer the existence of a hidden common cause (confounder) of a set of observed variables and reconstruct this confounder. The remainder of the paper is organized as follows: in Section 2, a method is proposed to determine (a), i.e., the number of mixture components $m$, Section 3 discusses established results on the identifiability of the parameters in (b). Section 4 presents an algorithm for determining these parameters and Section 5 uses the findings of the previous sections for confounder identification. Finally, the experiments are provided in Section 6.

## 2 Identifying the Number of Mixture Components

Various methods have been proposed in the literature to select the number of mixture components in a mixture model (e.g., Feng & McCulloch (1996); Böhning & Seidel (2003); Rasmussen (2000); Iwata et al. (2013)). However, they impose different kind of assumptions than the conditional independence assumption of model (1) (e.g., that the distributions of the components belong to a certain parametric family). Assuming model (1), Kasahara & Shimotsu (2010) proposed a nonparametric method that requires discretization of the observed variables and provides only a lower bound on $m$. In the following, we present a method to determine $m$ in (1) without making parametric assumptions on the component distributions.

### 2.1 Hilbert Space Embedding of Distributions

Our method relies on representing $P(X_1, \ldots, X_d)$ as a vector in a reproducing kernel Hilbert space (RKHS).

We briefly introduce this framework. For a random variable $X$ with domain $\mathcal{X}$, an RKHS $\mathcal{H}$ on $\mathcal{X}$ with kernel $k$ is a space of functions $f : \mathcal{X} \to \mathbb{R}$ with dot product $\langle \cdot, \cdot \rangle$, satisfying the reproducing property (Schölkopf & Smola, 2002): $\langle f(\cdot), k(x, \cdot) \rangle = f(x)$, and consequently, $\langle k(x, \cdot), k(x', \cdot) \rangle = k(x, x')$. The kernel thus defines a map $x \mapsto \phi(x) := k(x, .) \in \mathcal{H}$ satisfying $k(x, x') = \langle \phi(x), \phi(x') \rangle$, i.e., it corresponds to a dot product in $\mathcal{H}$.

Let $\mathcal{P}$ denote the set of probability distributions on $\mathcal{X}$, then we use the following *mean map* (Smola et al., 2007) to define a Hilbert space embedding of $\mathcal{P}$:

$$\mu : \mathcal{P} \to \mathcal{H}; \qquad P(X) \mapsto \mathbb{E}_X[\phi(X)] \tag{2}$$

We will henceforth assume this mapping to be injective, which is the case if $k$ is *characteristic* (Fukumizu et al., 2008), as the widely used Gaussian RBF kernel $k(x, x') = \exp(-\|x - x'\|^2 / (2\sigma^2))$.

We use the above framework to define Hilbert space embeddings of distributions of every single $X_j$. To this end, we define kernels $k_j$ for each $X_j$, with feature maps $x_j \mapsto \phi_j(x_j) = k(x_j, .) \in \mathcal{H}_j$. We thus obtain an embedding $\mu_j$ of the set $\mathcal{P}_j$ into $\mathcal{H}_j$ as in (2).

We can apply the same framework to embed the set of joint distributions $\mathcal{P}_{1,\ldots,d}$ on $\mathcal{X}_1 \times \ldots \times \mathcal{X}_d$. We simply define a joint kernel $k_{1,\ldots,d}$ by $k_{1,\ldots,d}((x_1, \ldots, x_d), (x'_1, \ldots, x'_d)) = \prod_{j=1}^d k_j(x_j, x'_j)$, leading to a feature map into $\mathcal{H}_{1,\ldots,d} := \bigotimes_{j=1}^d \mathcal{H}_j$ with $\phi_{1,\ldots,d}(x_1, \ldots, x_d) = \bigotimes_{j=1}^d \phi_j(x_j)$ (where $\bigotimes$ stands for the tensor product). We use the following mapping of the joint distribution:

$$\mu_{1,\ldots,d} : \mathcal{P}_{1,\ldots,d} \to \bigotimes_{j=1}^d \mathcal{H}_j \tag{3}$$

$$P(X_1, \ldots, X_d) \mapsto \mathbb{E}_{X_1,\ldots,X_d}[\bigotimes_{j=1}^d \phi_j(X_j)]$$

### 2.2 Identifying the Number of Components from the Rank of the Joint Embedding

By linearity of the maps $\mu_{1,\ldots,d}$ and $\mu_j$, the embedding of the joint distribution decomposes into

$$\begin{aligned} \mathcal{U}_{X_1,\ldots,X_d} :=& \mu_{1,\ldots,d}(P(X_1, \ldots, X_d)) \\ =& \sum_{i=1}^m P(z^{(i)}) \bigotimes_{j=1}^d \mathbb{E}_{X_j}[\phi_j(X_j) | z^{(i)}] \,. \end{aligned} \tag{4}$$

**Definition 1 (full rank conditional)** *Let $A, B$ two random variables with domains $\mathcal{A}, \mathcal{B}$, respectively. The conditional probability distribution $P(A|B)$ is called a full rank conditional if $\{P(A|b)\}_b$ with $b \in \mathcal{B}$ is a linearly independent set of distributions.*

Recalling that the rank of a tensor is the minimum number of rank 1 tensors needed to express it as a linear combination of them, we obtain:

**Theorem 1 (number of mixture components)** *If $P(X_1, \ldots, X_d)$ is decomposable as in (1) and $P(X_j|Z)$ is a full rank conditional for all $1 \leq j \leq d$, then the tensor rank of $\mathcal{U}_{X_1,\ldots,X_d}$ is $m$.*

**Proof.** From (4), the tensor rank of $\mathcal{U}_{X_1,\ldots,X_d}$ is at most $m$. If the rank is $m' < m$, there exists another decomposition of $\mathcal{U}_{X_1,\ldots,X_d}$ (apart from (4)) as $\sum_{i=1}^{m'} \bigotimes_{j=1}^d v_{i,j}$, with non-zero vectors $v_{i,j} \in \mathcal{H}_j$. Then, there exists a vector $w \in \mathcal{H}_1$, s.t. $w \perp \text{span}\{v_{1,1}, \ldots, v_{m',1}\}$ and $w \not\perp \text{span}\{(\mathbb{E}_{X_1}[\phi_1(X_1)|z^{(i)}])_{1\leq i \leq m}\}$. The dual vector $, w\rangle$ defines a linear form $\mathcal{H}_1 \to \mathbb{R}$. By overloading notation, we consider it at the same time as a linear map $\mathcal{H}_1 \otimes \cdots \otimes \mathcal{H}_d \to \mathcal{H}_2 \otimes \cdots \otimes \mathcal{H}_d$, by extending it with the identity map on $\mathcal{H}_2 \otimes \cdots \otimes \mathcal{H}_d$. Then, $\langle \sum_{i=1}^{m'} \bigotimes_{j=1}^d v_{i,j}, w \rangle = \sum_{i=1}^{m'} \langle v_{i1}, w \rangle \bigotimes_{j=2}^d v_{i,j} = 0$ but $\langle \mathcal{U}_{X_1,\ldots,X_d}, w \rangle \neq 0$. So, $m = m'$. □

The assumption that $P(X_j|Z)$ is a full rank conditional, i.e., that $\{P(X_j|z^{(i)})\}_{i \leq m}$ is a linearly independent set, is also used by Allman et al. (2009) (see Sec. 3). It does not prevent $P(X_j|z^{(q)})$ from being itself a mixture distribution, however, it implies that, for all $j, q$, $P(X_j|z^{(q)})$ is not a linear combination of $\{P(X_j|z^{(r)})\}_{r \neq q}$. Theorem 1 states that, under this assumption, the number of mixture components $m$ of (1) (or equivalently the number of values of $Z$) is identifiable and equal to the rank of $\mathcal{U}_{X_1,\ldots,X_d}$. A straightforward extension of Theorem 1 reads:

**Lemma 1 (infinite Z)** *If $Z$ takes values from an infinite set, then the tensor rank of $\mathcal{U}_{X_1,\ldots,X_d}$ is infinite.*

### 2.3 Empirical Estimation of the Tensor Rank from Finite Data

Given empirical data for every $X_j$, $\{x_j^{(1)}, x_j^{(2)}, \ldots, x_j^{(n)}\}$, to estimate the rank of (4), we replace it with the empirical average

$$\hat{\mathcal{U}}_{X_1,\ldots,X_d} := \frac{1}{n} \sum_{i=1}^n \bigotimes_{j=1}^d \phi_j(x_j^{(i)}) \,, \tag{5}$$

which is known to converge to the expectation in Hilbert space norm (Smola et al., 2007).

The vector (5) still lives in the infinite dimensional feature space $\mathcal{H}_{1,\ldots,d}$, which is a space of functions

$\mathcal{X}_1 \times \cdots \times \mathcal{X}_d \to \mathbb{R}$. To obtain a vector in a finite dimensional space, we evaluate this function at the $n^d$ data points $(x_1^{(q_1)}, \ldots, x_d^{(q_d)})$ with $q_j \in \{1, \ldots, n\}$ (the $d$-tuple of superscripts $(q_1, \ldots, q_d)$ runs over all elements of $\{1, \ldots, n\}^d$). Due to the reproducing kernel property, this is equivalent to computing the inner product with the images of these points under $\phi_{1,\ldots,d}$:

$$V_{q_1,\ldots,q_d} := \left\langle \hat{\mathcal{U}}_{X_1,\ldots,X_d}, \bigotimes_{j=1}^{d} \phi_j(x_j^{(q_j)}) \right\rangle = \frac{1}{n} \sum_{i=1}^{n} \prod_{j=1}^{d} k_j(x_j^{(i)}, x_j^{(q_j)}) \quad (6)$$

For $d = 2$, $V$ is a matrix, so one can easily find low rank approximations via truncated Singular Value Decomposition (SVD) by dropping low SVs. For $d > 2$, finding a low-rank approximation of a tensor is an ill-posed problem (De Silva & Lim, 2008). By grouping the variables into two sets, say $X_1, \ldots, X_s$ and $X_{s+1}, \ldots, X_d$ without loss of generality, we can formally obtain the $d = 2$ case with two vector-valued variables. This amounts to reducing $V$ in (6) to an $n \times n$ matrix by setting $q_1 = \cdots = q_s$ and $q_{s+1} = \cdots = q_d$. In theory, we expect the rank to be the same for all possible groupings. In practice, we report the rank estimation of the majority of all groupings. The computational complexity of this step is $O(2^{d-1}N^3)$.

# 3 Identifiability of Component Distributions and Mixing Weights

Once we have determined the number of mixture components $m$, we proceed to step (b) of recovering the distribution of each component $P(X_1, \ldots, X_d|z^{(i)})$ and the mixing weights $P(z^{(i)})$. In the following, we describe results from the literature on when these parameters are identifiable, for known $m$. Hall & Zhou (2003) proved that when $m = 2$, identifiability of parameters always holds in $d \geq 3$ dimensions. For $d = 2$ and $m = 2$ the parameters are generally not identifiable (there is a two-parameter continuum of solutions). In this case, one can obtain identifiability if, for all $j$, $P(X_j|Z)$ is pure (a conditional $P(X|Y)$ is called pure if for any two values $y, y'$ of $Y$, the sum $P(X|y)\lambda + P(X|y')(1 - \lambda)$ is a probability distribution only for $\lambda \in [0, 1]$ (Janzing et al., 2011)). Allman et al. (2009) established identifiability of the parameters whenever $d \geq 3$ and for all $m$ under weak conditions [1], using a theorem of Kruskal (1977). Finally, Kasahara & Shimotsu (2010) provided complementary identifiability results for $d \geq 3$ under different conditions with a constructive proof.

[1] the same assumption used in Theorem 1 that $P(X_j|Z)$ is a full rank conditional for all $j$.

# 4 Identifying Component Distributions and Mixing Weights

We are given $n$ data points from an identifiable distribution $P(X_1, \ldots, X_d)$. Our goal is to cluster the data points (using $m$ labels) in such a way that the distribution of points with label $i$ is close to the (unobserved) empirical distribution of every mixture component, $P_n(X_1, \ldots, X_d|z^{(i)})$.

## 4.1 Existing Methods

Probabilistic mixture models or other clustering methods can be used to identify the mixture components (clusters) (e.g., Von Luxburg (2007); Böhning & Seidel (2003); Rasmussen (2000); Iwata et al. (2013)). However, they impose different kind of assumptions than the conditional independence assumption of model (1) (e.g., Gaussian mixture model). Assuming model (1), Levine et al. (2011) proposed an Expectation-Maximization (EM) algorithm for nonparametric estimation of the parameters in (1), given that $m$ is known. Their algorithm uses a kernel as smoothing operator. They choose a common kernel bandwidth for all the components because otherwise their iterative algorithm is not guaranteed not to increase from one iteration to another. As stated also by Chauveau et al. (2010), the fact that they do not use an adaptive bandwidth (Benaglia et al., 2011) can be problematic especially when the distributions of the components differ significantly.

## 4.2 Proposed Method: Clustering with Independence Criterion (CLIC)

The proposed method, CLIC, assigns each of the $n$ observations to one of the $m$ (as estimated in Section 2) mixture components (clusters). We do not claim that each single data point is assigned correctly (especially when the clusters are overlapping). Instead, we aim to yield the variables jointly conditionally independent given the cluster in order to recover the components.

To measure conditional independence of $X_1, \ldots, X_d$ given the cluster we use the Hilbert Schmidt Independence Criterion (HSIC) (Gretton et al., 2008). It measures the Hilbert space distance between the kernel embeddings of the joint distribution of two (possibly multivariate) random variables and the product of their marginal distributions. If $d > 2$, we test for mutual independence. For that, we perform multiple tests, namely: $X_1$ against $(X_2, \ldots, X_d)$, then $X_2$ against $(X_3, \ldots, X_d)$ etc. and use Bonferroni correction. For each cluster, we consider as test statistic the HSIC from the test that leads to the smallest $p$-value (“highest” dependence).

We regard the negative sum of the logarithms of all $p$-values (each one corresponding to one cluster) under the null hypothesis of independence as our objective function. The proposed algorithm is iterative. We first randomly assign every data point to one mixture component. In every iteration we perform a greedy search: we randomly divide the data into disjoint sets of $c$ points. Then, we select one of these sets and consider all possible assignments of the set's points to the $m$ clusters ($m^c$ possible assignments). The assignment that optimizes the objective function is accepted and the points of the set are assigned to their new clusters (which may coincide with the old ones). We, eventually, repeat the same procedure for all disjoint sets and this constitutes one iteration of our algorithm. After every iteration we test for conditional independence given the cluster. The algorithm stops after an iteration when any of the following happens: we observe independence in all clusters, no data point has changed cluster assignment, an upper limit of iterations is reached. It is clear that the objective function is monotonously decreasing.

The algorithm may not succeed at producing conditionally independent variables for different reasons: e.g., incorrect estimation of $m$ from the previous step or convergence to a local optimum. In that case, CLIC reports that it was unable to find appropriate clusters.

Along the iterations, the kernel test of independence updates the bandwidth according to the data points belonging to the current cluster (in every dimension). Note, however, that this is not the case for the algorithm in Section 2. There, we are obliged to use a common bandwidth, because we do not have yet any information about the mixture components.

The parameter $c$ allows for a trade-off between speed and avoiding local optima: for $c = n$, CLIC would find the global optimum after one step, but this would require checking $m^n$ cluster assignments. On the other hand, $c = 1$ leads to a faster algorithm that may get stuck in local optima. In all experiments we used $c = 1$ since we did not encounter many problems with local optima. Considering $c$ to be a constant, the computational complexity of CLIC is $O(m^c N^3)$ for every iteration. Algorithm 1 includes the pseudocode of CLIC.

**Algorithm 1** CLIC

**input** data matrix $X$ of size $n \times d$, $m$, $c$
random assignment $cluster(i) \in \{1, \ldots, m\}, i = 1, \ldots, n$ of the data into $m$ clusters
**while** conditional dependence given cluster and clusters change **do**
  $obj = computeObj(cluster)$
  choose random partition $S_j, j = 1, \ldots, J$ of the data into sets of size $c$
  **for** $j = 1$ **to** $J$ **do**
    $newCluster = cluster$
    **for all** words $w \in \{1, \ldots, m\}^c$ **do**
      $newCluster(S_j) = w$
      $objNew(w) = computeObj(newCluster)$
    **end for**
    $wOpt = \text{argmin}(objNew)$
    $cluster(S_j) = wOpt$
  **end for**
**end while**
**if** conditional independence given cluster **then**
  **output** $cluster$
**else**
  **output** "Unable to find appropriate clusters."
**end if**

## 5 Causality: Identifying Confounders

Drawing causal conclusions from observed statistical dependences without being able to intervene on the system always relies on assumptions that link statistics to causality (Spirtes et al., 2001; Pearl, 2000). The least disputed one is the Causal Markov Condition (CMC) stating that every variable is conditionally independent of its non-descendant, given its parents (Spirtes et al., 2001; Pearl, 2000) with respect to a directed acyclic graph (DAG) that formalizes the causal relations. We focus on causal inference problems where the set of observed variables may not be causally sufficient, i.e. statistical dependences can also be due hidden common causes (confounders) of two or more observed variables. Under the assumption of linear relationships between variables and non-Gaussian distributions, confounders may be identified using Independent Component Analysis (Shimizu et al., 2009). Other results for the linear case are presented in Silva et al. (2006) and for the non-linear case with additive noise in Janzing et al. (2009). Fast Causal Inference (Silva et al., 2006) can exclude confounding for some pairs of variables, given that many variables are observed. Finally, the reconstruction of binary confounders under the assumption of pure conditionals is presented in Janzing et al. (2011).

In this section, we use the results of the previous sections to infer the existence and identify a finite confounder that explains all the dependences between the observed variables. We assume that latent variables are not caused by observed variables (the same assumption has been used by Silva et al. (2006)). Unlike previous methods, we do not make explicit assumptions on the distribution of the variables. Instead, we postulate a different assumption, namely that the conditional of each variable given its parents is full rank:

**Assumption 1 (full rank given parents)** *Let $PA_Y$ denote the parents of a continuous variable $Y$ with respect to the true causal DAG. Then, $P(Y|PA_Y)$ is a full rank (f.r.) conditional.*

**Lemma 2 (full rank given parent)** *By Assumption 1 it follows that, if $A \in PA_Y$ is one of the parents of $Y$, i.e., $A \rightarrow Y$, then, since $P(Y|PA_Y)$ is a f.r. conditional, $P(Y|A)$ is also a f.r. conditional (after marginalization).*

Remark: If Assumption 1 holds and $A \rightarrow B \rightarrow C$, then $P(B|A)$ and $P(C|B)$ are f.r. conditionals (Lemma 2), which implies that $P(C|A)$ is also a f.r. conditional, since it results from their multiplication.

**Lemma 3 (shifted copies)** *Let $R$ be a probability distribution on $\mathbb{R}$ and $T_t R$ its copy shifted by $t \in \mathbb{R}$ to the right. Then $\{T_t R\}_{t \in \mathbb{R}}$ are linearly independent.*

**Proof.** Let

$$\sum_{j=1}^{q} \alpha_j T_{t_j} R = 0 \,, \tag{7}$$

for some $q$ and some $q$-tuple $\alpha_1, \ldots, \alpha_q$. Let $\hat{R}$ be the Fourier transform of $R$. If we set $g(\omega) := \sum_{j=1}^{q} \alpha_j e^{i\omega t_j}$ then (7) implies $g(\omega)\hat{R}(\omega) = 0$ for all $\omega \in \mathbb{R}$, hence $g$ vanishes for all $\omega$ with $\hat{R}(\omega) \neq 0$, which is a set of non-zero measure. Since $g$ is holomorphic, it therefore vanishes for all $\omega \in \mathbb{R}$ and thus all coefficients are zero. □

The following theorem shows that Assumption 1 is typically satisfied for a class of causal models considered by Hoyer et al. (2009):

**Theorem 2 (additive noise model)** *Let $Y$ be given by $Y = f(PA_Y) + N$, where $N$ is a noise variable that is statistically independent of $PA_Y$ and $f$ is an injective function. Then $P(Y|PA_Y)$ is a full rank conditional.*

The proof is a straightforward application of Lemma 3.

**Theorem 3 (rank of cause-effect pair)** *Assume there is a direct causal link $A \rightarrow B$ between two observed variables $A$ and $B$, with $\{A,B\}$ not necessarily being a causally sufficient set (i.e., there may be unobserved common causes of $A$ and $B$). Then, given Assumption 1, the rank of $\mathcal{U}_{A,B}$ is equal to the number of values that $A$ takes, if $A$ is finite. If $A$ is infinite, then the rank of $\mathcal{U}_{A,B}$ is infinite.*

**Proof.** By Assumption 1, $P(B|A)$ is a f.r. conditional (Lemma 2). Since $A \perp\!\!\!\perp B \,|\, A$, applying Theorem 1 for finite $Z := A$ we conclude that the rank of $\mathcal{U}_{A,B}$ is equal to the number of values of $A$. For infinite $A$, we similarly apply Lemma 1 and we get infinite rank of $\mathcal{U}_{A,B}$. □

**Theorem 4 (rank of confounded pair)** *Assume $A \leftarrow C \rightarrow B$, and $A$, $B$ do not have other common causes apart from $C$. Then, given Assumption 1, the rank of $\mathcal{U}_{A,B}$ is equal to the number of values of $C$.*

**Proof.** The proof is straightforward: by Assumption 1, $P(A|C)$ and $P(B|C)$ are f.r. conditionals (Lemma 2). Additionally, $A \perp\!\!\!\perp B \,|\, C$ and then, according to Theorem 1, the rank of $\mathcal{U}_{A,B}$ is equal to the number of values of $Z := C$ (for infinite $C$, the rank is infinite). □

Theorems 3 and 4 state what is the expected rank of $\mathcal{U}_{A,B}$ for various causal structure scenarios. However, in causal inference we are interested in inferring the *unknown* underlying causal DAG. The following Theorem uses Theorem 3 to infer the causal structure based on the rank of the embedding of the joint distribution.

**Theorem 5 (identifying confounders)** *Let $Y_1, \ldots, Y_d$ denote all observed variables. Assume they are continuous, pairwise dependent, and there is at most one (if any) hidden common cause of two or more of the observed variables. If Assumption 1 holds and the rank of $\mathcal{U}_{Y_1,\ldots,Y_d}$, with $d \geq 3$, is finite, then Fig. 1 depicts the only possible causal DAG with $W$ being an unobserved variable and $P(Y_1, \ldots, Y_d, W)$ is identifiable up to reparameterizations of $W$.*

**Proof.** Assume there is at least one direct causal link $Y_i \rightarrow Y_{i'}$. Then, according to Theorem 3, the rank of $\mathcal{U}_{Y_i,Y_{i'}}$, and thus the rank of $\mathcal{U}_{Y_1,\ldots,Y_d}$, would be infinite. Therefore, direct causal links between the $\{Y_j\}$ can be excluded and the statistical dependencies between $\{Y_j\}$ can only be explained by hidden common causes. Since we assumed that there is at most one hidden common cause (and the observed variables are pairwise dependent), the only possible causal graph is depicted in Fig. 1. This implies $Y_1 \perp\!\!\!\perp Y_2 \perp\!\!\!\perp \ldots \perp\!\!\!\perp Y_d \,|\, W$ (according to the CMC), so model (1) holds, with $Z := W$ being the latent variable. According to the previous sections (Theorem 1 and Section 3), this model is identifiable. □

Based on Theorem 1, the number of values of $W$ is equal to the rank of $\mathcal{U}_{Y_1,\ldots,Y_d}$ and $P(Y_1, \ldots, Y_d, W)$ can be identified according to Section 4. Note that the single common cause $W$ could be the result of merging many common causes $W_1, .., W_k$ to one vector-valued variable $W$. Thus, at first glance, it seems that one does not lose generality by assuming only one common cause. However, Assumption 1, then, excludes the case where $W$ consists of components each of which only acts on some different subset of the $\{Y_j\}$. $W_1, .., W_k$ should all be common causes of all $\{Y_j\}$.

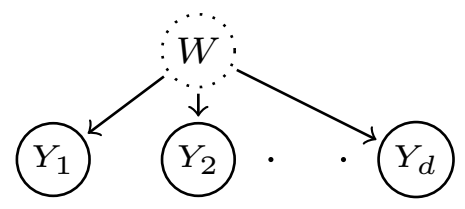

Figure 1: Inferred causal DAG (the dotted circle represents an unobserved variable).

Note that, when we are given only finite data, the estimated rank of $\mathcal{U}_{Y_1,\ldots,Y_d}$ is always finite, highly depending on the strength of the causal arrows and the sample size. Then, we are faced with the issue that, based on Theorem 5, we would always infer that Fig. 1 depicts the only possible causal DAG, with the number of values of $W$ being equal to the estimated rank. However, the lower the rank, the more confident we get that this is, indeed, due to the existence of a confounder that renders the observed variables conditionally independent (Fig. 1). On the other hand, high rank can also be due to direct causal links between the observed variables or continuous confounders. For that, we consider Theorem 5 to be more appropriate for inferring the existence of a confounder with a small number of values which would lead to low rank. However, we admit that what is considered "high" or "low" is not well defined. For example, how much "high" rank values we expect for the DAG $Y_1 \to Y_2$ highly depends on the strength of the causal arrow: the lower the dependence between $Y_1$ and $Y_2$ is, the lower is, generally, the estimated rank. In practice, we could make a vague suggestion that whenever the estimated rank is below 5 (although the dependence between $\{Y_j\}$ is strong), it is quite probable that this is due to a confounder (Fig. 1) but for higher rank it is getting more difficult to decide on the underlying causal structure.

# 6 Experiments

We conducted experiments both on simulated and real data. In all our experiments we use a Gaussian RBF kernel $k(x, x') = \exp\left(-\|x - x'\|^2/(2\sigma^2)\right)$. Concerning the first step of determining the number of mixture components: a common way to select the bandwidth $\sigma_j$ for every $k_j$ is to set it to the median distance between all data points in the $j$th dimension of the empirical data. However, this approach would usually result in an overestimation of the bandwidth, especially in case of many mixture components (see also Benaglia et al. (2011)). To partially account for this, we compute the bandwidth for every $X_j$ as the median distance between points in the neighborhood of every point in the sample. The neighborhood is found by the 10 nearest neighbors of each point computed using all other variables apart from $X_j$. To estimate the rank of $V$, we find its SVD and report the estimated rank as $\hat{m} = \mathrm{argmin}_i(SV_{i+1}/SV_i)$ within the SVs that cover 90-99.999% of the total variance. Finally, concerning CLIC, we use 7 as the maximum number of iterations, but usually the algorithm terminated earlier.

## 6.1 Simulated Data

Simulated data were generated according to the DAG of Fig. 1 (we henceforth refer to them as the first set of simulated data), e.i., model (1) holds with $Z := W$, since $Y_1 \perp\!\!\!\perp Y_2 \perp\!\!\!\perp \ldots \perp\!\!\!\perp Y_d \,|\, W$. We first generated $Z$ from a uniform distribution on $m$ values. Then, the distribution of each mixture component in every dimension ($P(X_j|z^{(i)})$) was chosen randomly between: (i) a normal distribution with standard deviation 0.7, 1, or 1.3, (ii) a t-distribution with degrees of freedom 3 or 10, (iii) a (stretched) beta distribution with alpha 0.5 or 1 and beta 0.5 or 1, and (iv) a mixture of two normal distributions with variance 0.7 for each. The distance between the components in each dimension was distributed according to a Gaussian with mean 2 and standard deviation 0.3. We chose the distance and the mixtures such that the experiments cover different levels of overlap between the components, and at the same time $\{P(X_j|z^{(i)})\}_{i\leq m}$ are generically linearly independent. We ran 100 experiments for each combination of $d = 2, 3, 5$ and $m = 2, 3, 4, 5$, with the sample size being $300 \times m$.

For comparison, we additionally generated data where the observed variables are connected also with direct causal links and thus are conditionally dependent given the confounder (we henceforth refer to them as the second set of simulated data). For that, we first generated data according to the DAG of Fig. 1, as above, for $d = 2$ and $m = 1$ (which amounts to no confounder) and for $d = 2$ and $m = 3$ (3-state confounder). $X_2$ was then shifted by $4X_1$ to simulate a direct causal link $X_1 \to X_2$. In this case, $X_1 \perp\!\!\!\perp X_2 \,|\, X_1$, so we have infinite $Z := X_1$.

### 6.1.1 Identifying the Number of Mixture Components

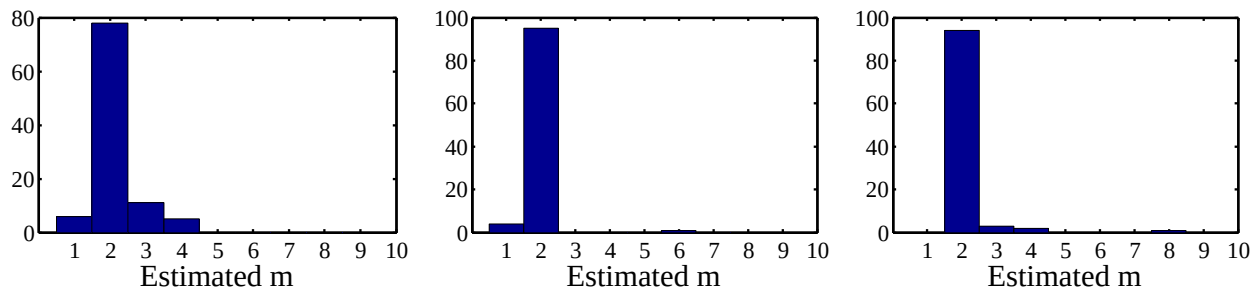

Figure 2: Histograms of the estimated number of mixture components $m$ for the first set of simulated data, for $m = 2$ throughout, and $d = 2$ (left), $d = 3$ (middle), $d = 5$ (right).

We first report results on the first part of identifica-tion, i.e. identifying the number of mixture components $m$ in (1). The empirical rank estimation may depend on the strength of the causal arrows, the kernel bandwidth selection, the sample size and the way to estimate the rank by keeping only large eigenvalues. Figures 2, 3, 4 and 5 illustrate histograms of the estimated number of components (equivalently the estimated number of values of the confounder) for the first set of simulated data for $m = 2, 3, 4$ and 5, respectively. Each figure contains one histogram for every value of $d = 2, 3$ and 5. We can observe that as $m$ increases the method becomes more sensitive in underestimating the number of components, a behavior which can be explained by the common sigma selection for all the data in each dimension or by high overlap of the distributions (which could violate Assumption 1). On the other hand, as $d$ increases the method becomes more robust in estimating $m$ correctly due to the grouping of variables that allows multiple rank estimations. The “low” estimated rank values provide us with some evidence that the causal DAG of Fig. 1 is true (Theorem 5). Of course, as stated also at the end of Sec. 5, it is difficult to define what is considered a low rank.

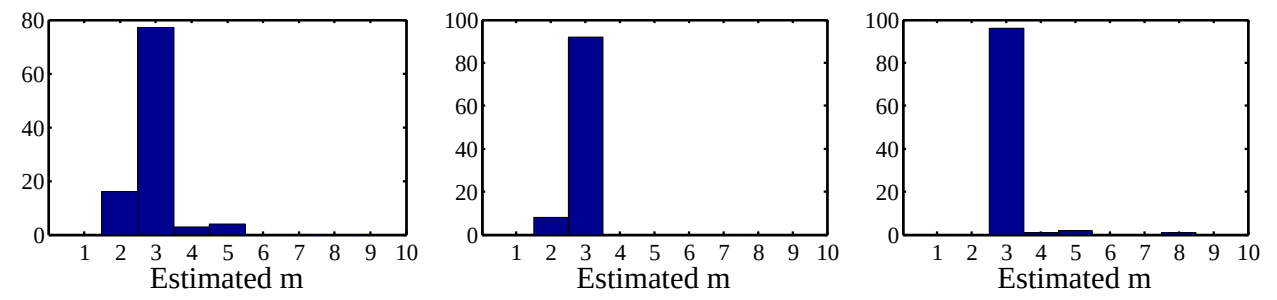

Figure 3: As Figure 2 but for $m = 3$.

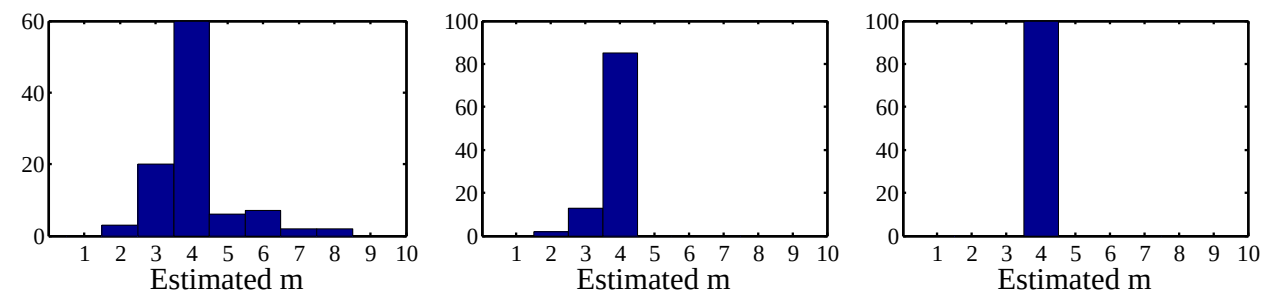

Figure 4: As Figure 2 but for $m = 4$.

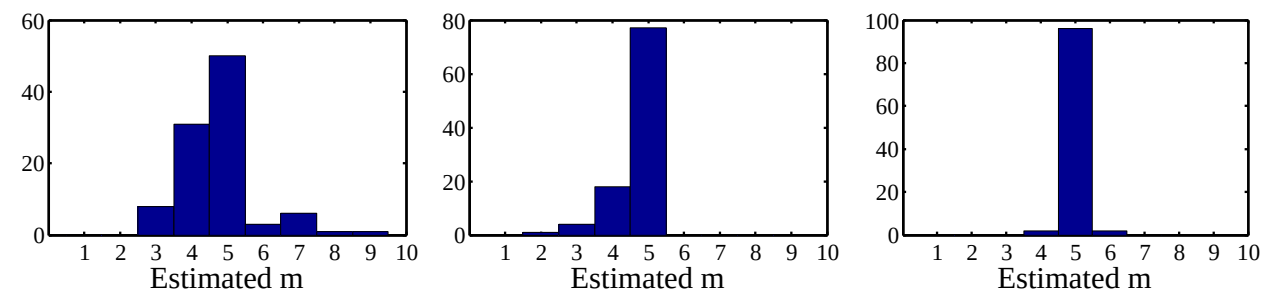

Figure 5: As Figure 2 but for $m = 5$.

Figure 6 depicts histograms of the estimated number of components for the second set of simulated data. According to Theorem 3, the direct causal link $X_1 \to X_2$ results in an infinite rank of $\mathcal{U}_{X_1,X_2}$. Indeed, we can observe that in this case the estimated $m$ is much higher. The “high” estimated rank values provide us with some evidence that the underlying causal DAG may include direct causal links between the observed variables or confounders with a high or infinite number of values. Note that, depending on the strength of the causal arrow $X_1 \to X_2$, we may get higher or lower rank values. For example, if the strength is very weak we get lower rank values since the dependence between $X_1$ and $X_2$ tends to be dominated by the confounder (that has a small number of values).

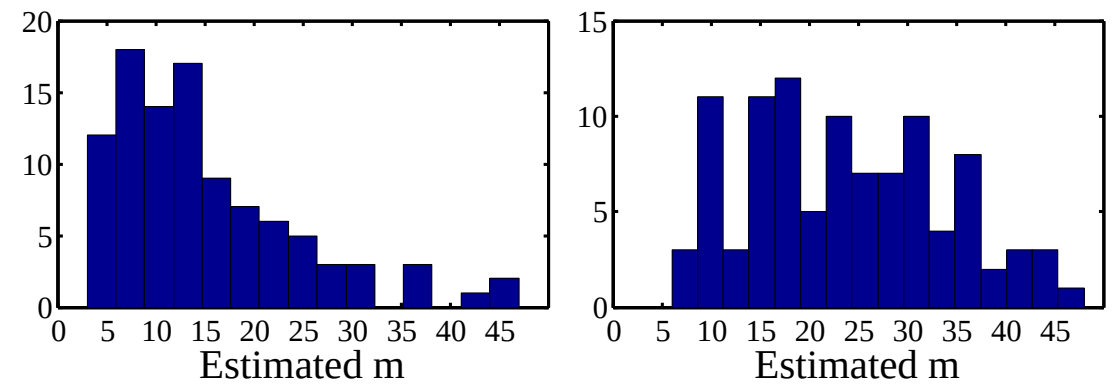

Figure 6: Histograms of the estimated $m$ for the second set of simulated data (including a direct causal arrow). Left: no confounder, right: 3-state confounder.

#### 6.1.2 Full Identification Framework

Next, we performed experiments, using the first set of simulated data to evaluate the performance of the proposed method (CLIC) (Section 4.2), the method of Levine et al. (2011) (Section 4.1) and the EM algorithm using a Gaussian mixture model (EM is repeated 5 times and the solution with the largest likelihood is reported). In the following, we refer to these methods as CLIC, Levine, and EM, respectively. For each data point, the two latter methods output posterior probabilities for the $m$ clusters, which we sample from to obtain cluster assignments. Figure 7 illustrates the cluster assignments of these three methods for one simulated dataset with $m = 3$ and $d = 2$ (this is more for visualization purposes because for these values of $d$ and $m$ model (1) is not always identifiable). Note that permutations of the colors are due to the ambiguity of labels in the identification problem. However, EM incorrectly identifies a single component (having a mixture of two Gaussians as marginal density in $X_1$ dimension) as two distinct components. It is clear that this is because it assumes that the data are generated by a Gaussian mixture model and not by model (1), as opposed to CLIC and Levine methods.

We compare the distribution of each cluster output (for each of the three methods) to the empirical distribution, $P_n(X_1, \ldots, X_d | z^{(i)})$, of the corresponding mixture component (ground truth). For that we use the squared maximum mean discrepancy (MMD) (Gretton et al., 2012) that is the distance between Hilbert space embeddings of distributions. We only use the MMD and not one of the test statistics described in Gretton et al. (2012), since they are designed to compare two independent samples, whereas our samples

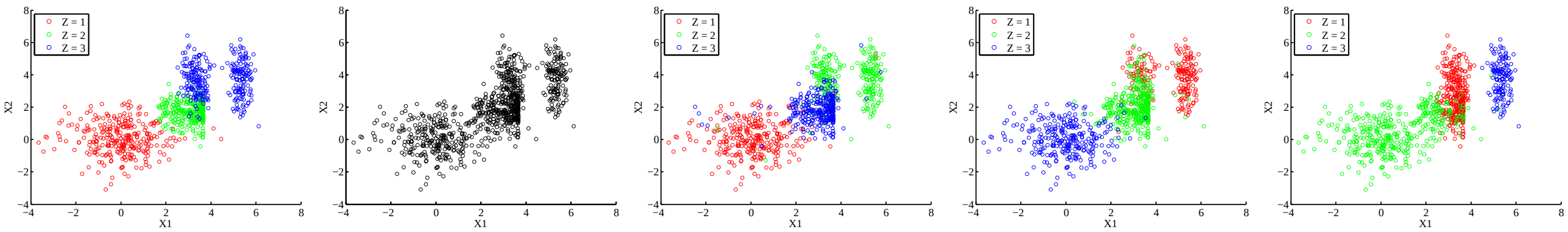

Figure 7: From left to right: ground truth, input, CLIC output, Levine output, and EM output for simulated data generated for $m = 3$. Each color represents one mixture component.

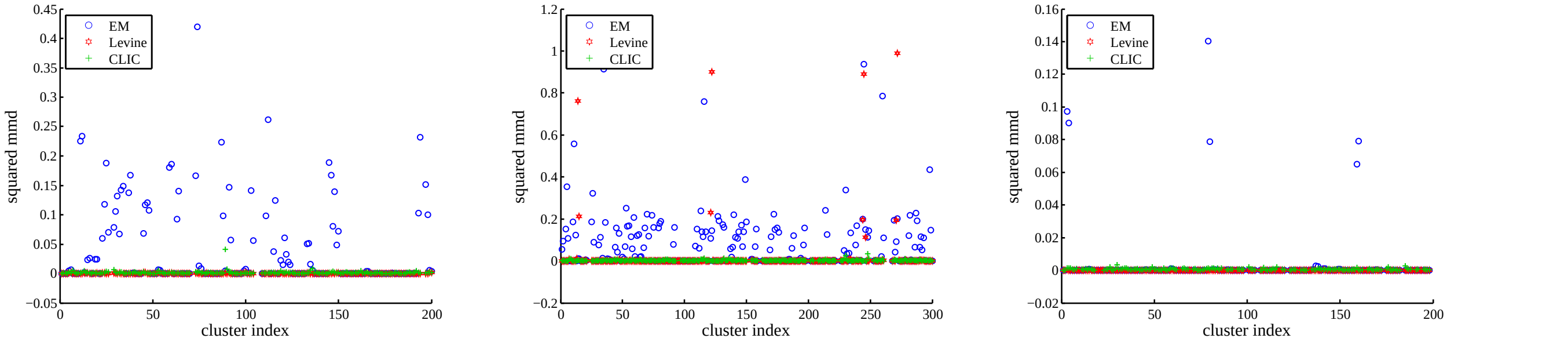

(a) (b) (c)

Figure 8: Squared MMD between output and ground truth clusters (for each of the three methods) for simulated data with (a) $d = 3, m = 2$, (b) $d = 3, m = 3$ and (c) $d = 5, m = 2$

(output and ground truth) have overlapping observations. To account for the permutations, we measure the MMD for all cluster permutations and select the one with the minimum sum of MMD for all clusters. Figures 8(a)-8(c) report the squared MMD results of the three methods for different combinations of $m$ and $d$. Each point corresponds to the squared MMD for one cluster of one of the 100 experiments. Results are provided only for the cases that the number of components $m$ was correctly identified from the previous step. The CLIC method was unable to find appropriate clusters in 2 experiments for $d = 3$ and $m = 3$ and in 13 for $d = 5$ and $m = 2$. Without claiming that the comparison is exhaustive, we can infer that both CLIC and Levine methods perform significantly better than EM, since they impose conditional independence. For higher $d$, EM improves since the clusters are less overlapping. However, the computational time of CLIC is higher compared to the other two methods.

## 6.2 Real data

Further, we applied our framework to the Breast Cancer Wisconsin (Diagnostic) Data Set from the UCI Machine Learning Repository (Frank & Asuncion, 2010). The dataset consists of 32 features of breast masses along with their classification as benign (B) or malignant (M). The sample size of the dataset is 569 (357 (B), 212 (M)). We selected 3 features, namely perimeter, compactness and texture, which are pairwise dependent (the minimum p-value is $pval = 2.43e - 17$), but become (close to) mutually independent when we condition on the class (B or M) ($pval_B = 0.016, pval_M = 0.013$). We applied our method to these three features (assuming the class is unknown) and we succeeded at correctly inferring that the number of mixture components is 2. Figure 9 depicts the ground truth of the breast data, the input and the results of CLIC, Levine and EM, and Fig. 12(a) the corresponding squared MMDs. We can observe that Levine method performs very poorly for this dataset.

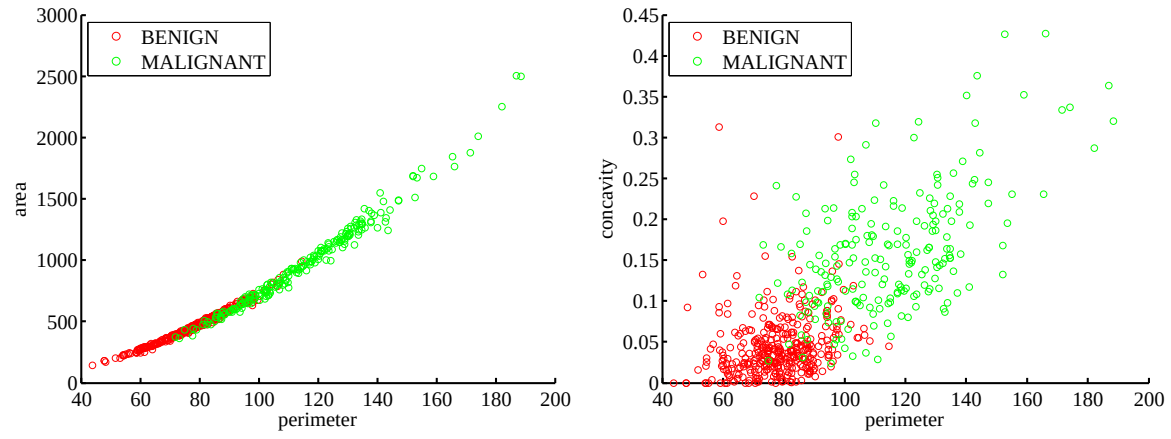

Figure 10: Breast data: features conditionally dependent given the class. Left: estimated $m = 62$, right: estimated $m = 8$.

Additionally, we selected different features, namely perimeter and area, and concavity and area, which are not conditionally independent given the binary class.

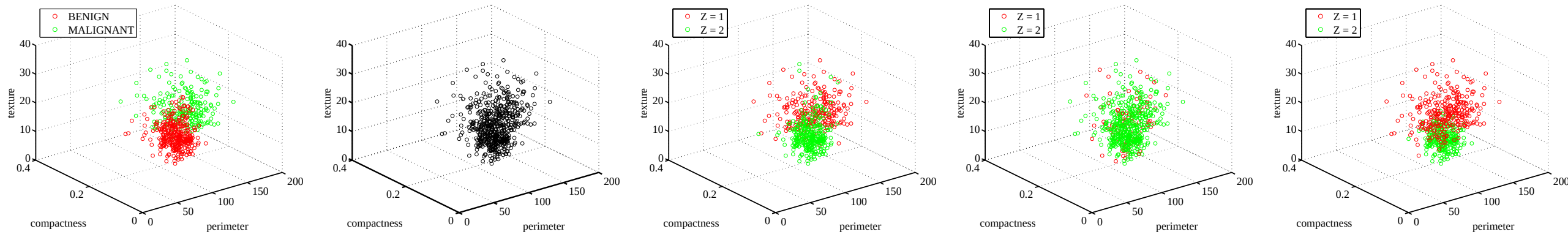

Figure 9: From left to right: ground truth, input, output CLIC, Levine, and EM for the breast data.

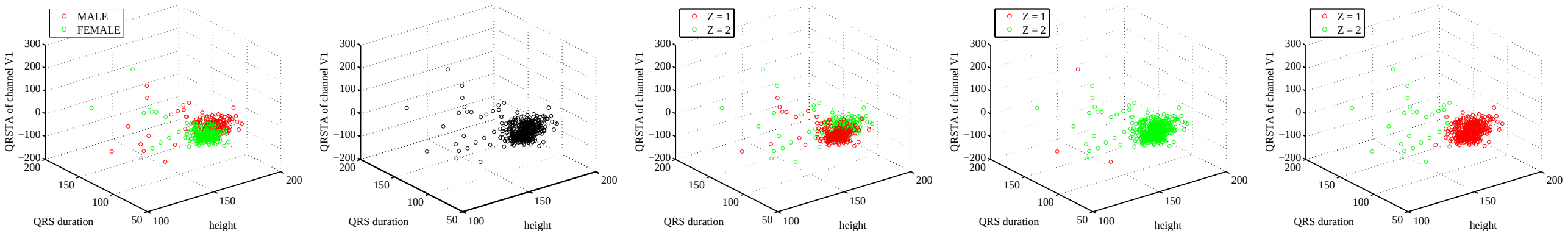

Figure 11: From left to right: ground truth, input, output CLIC, Levine, and EM for the arrhythmia data.

In this case, we got rank values higher than two, i.e., 62 and 8, respectively (Fig. 10).

We similarly applied our framework to the Arrhythmia Data Set (sample size 452) (Frank & Asuncion, 2010). We selected 3 features, namely height, QRS duration and QRSTA of channel V1 which are dependent (minimum $pval = 8.96e - 05$), but become independent when we condition on a fourth feature, the sex of a person (Male or Female) ($pval_M = 0.0607, pval_F = 0.0373$). We applied our method to the three features and we succeeded at correctly inferring that the number of mixture components is 2. Figure 11 depicts the ground truth, the input and the results of CLIC, Levine and EM, and Fig. 12(b) the corresponding squared MMDs. We can observe that Levine and EM methods perform very poorly for this dataset.

Finally, we applied our method to a database with cause-effect pairs[2]. It includes pairs of variables with known causal structure. Since there exists a direct causal arrow $X \to Y$, we expect the rank of $\mathcal{U}_{X,Y}$ to be infinite given our assumptions (Theorem 3), even if there exist hidden confounders or not. However, the estimated rank from finite data is always finite, its magnitude strongly depending on the strength of the causal arrow and the sample size, as mentioned in Sec. 5. Figure 13 depicts 4 cause-effect pairs (which were taken from the UCI Machine Learning Repository (Frank & Asuncion, 2010)) with the same sample size (1000 data points) but various degrees of dependence, specifically: $pval = 7.16e - 12$, $pval = 9.41e - 63$, $pval = 1.21e - 317$, $pval = 0$. The estimated ranks are $m = 1, 4, 8$ and 63, respectively. Note that when $X$ and $Y$ are close to independent (e.g., the first plot of Fig. 13) the assumption of pairwise dependence of Theorem 5 is almost violated.

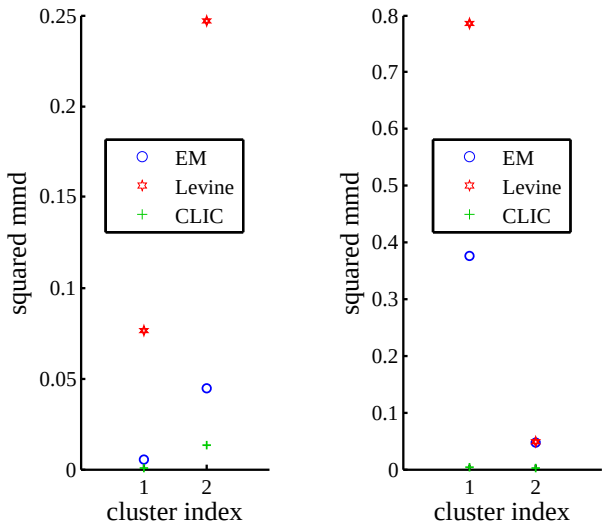

(a) (b)

Figure 12: Squared MMD between output and ground truth clusters for (a) breast and (b) arrhythmia data.

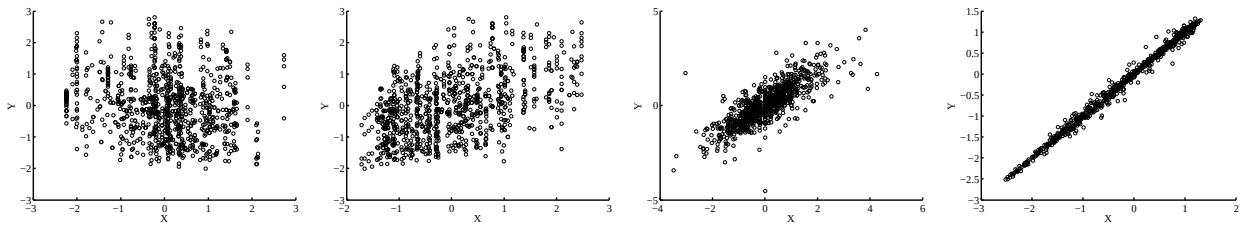

Figure 13: Four cause-effect pairs. Estimated $m$ from left to right: $m = 1$, $m = 4$, $m = 8$, and $m = 63$.

## 7 Conclusion

In this paper, we introduced a kernel method to identify finite mixtures of nonparametric product distributions. The method was further used to infer the existence and identify a hidden common cause of a set of observed variables. Experiments on simulated and real data were performed for evaluation of the proposed approach. In practice, our method is more appropriate for the identification of a confounder with a small number of values.

[2] http://webdav.tuebingen.mpg.de/cause-effect/